Original Paper

# Identifying Public Response Topics to CDC COVID-19 Communications on Social Media: Infoveillance Study Using Large Language Model–Based Rephrasing

Wangjiaxuan Xin[1]*, Shuhua Yin[2], Shi Chen[2], Yaorong Ge[1]

[1]College of Computing and Informatics, University of North Carolina at Charlotte, Charlotte, North Carolina, NC, United States
[2]Department of Epidemiology and Community Health, University of North Carolina at Charlotte, Charlotte, NC, United States

*Corresponding Author:
Wangjiaxuan Xin
University of North Carolina at Charlotte 9201 University City Blvd
Charlotte, NC, 28223
United States
Phone: +1 7048776536
Email: wxin@charlotte.edu

## Abstract

**Background:** Public health agencies increasingly rely on social media platforms such as X (formerly Twitter) to monitor public responses to health communications and to support timely, data-driven decision-making during health crises such as the COVID-19 pandemic. In particular, public replies to official communications from agencies such as the Centers for Disease Control and Prevention (CDC) provide valuable insights into population-level perceptions, concerns, and engagement with health policies. However, effectively analyzing these large-scale short-text responses remains challenging due to their brevity, informality, and linguistic variability. Topic modeling offers a scalable approach to identifying thematic patterns in such data, but its performance is often limited when applied to short, noisy social media text, resulting in low-quality and difficult-to-interpret topics. Although recent advances in large language models (LLMs) provide new opportunities to enhance text representation, their potential to improve topic modeling for public health–oriented social media analysis remains underexplored.

**Objective:** This study aims to enhance public health surveillance by improving the analysis of public responses to official health communications on social media. To achieve this, we develop and evaluate TM-Rephrase, a model-agnostic framework that leverages LLM–based rephrasing to improve the quality, interpretability, and semantic relevance of topics derived by topic models for public health-related short texts on social media.

**Methods:** We analyzed 25,027 public replies to official CDC posts on X collected between May 2020 and November 2022 to examine public responses to health communications. We applied a LLM–based rephrasing framework (TM-Rephrase) to transform informal short texts into more standardized and context-enriched representations using general and colloquial-to-formal schemes. Both original and rephrased texts were analyzed across multiple topic models, and topic quality was evaluated using coherence, uniqueness, redundancy, and diversity metrics, along with qualitative assessment of interpretability and post-topic semantic alignment.
**Results:** TM-Rephrase consistently improved topic quality across models and evaluation metrics. For LDA, topic coherence increased from $C_v$ =0.3094 (no rephrasing) to $C_v$ =0.5004 with colloquial-to-formal rephrasing (>60% relative improvement). For BERTopic, coherence improved from $C_v$=0.4078 to $C_v$ =0.4734. Diversity-related metrics also improved, with TSCTM achieving TU=1.0, TD=1.0, and TR=0 under rephrased conditions, indicating fully distinct topics. These improvements were robust across multiple LLMs (Gemini, GPT-4o-mini, and Mistral-7B). Qualitative results further showed that rephrased texts produced more interpretable and semantically coherent topics (representative keywords), enabling clearer identification of public concerns and responses to health communications, including themes related to vaccination attitudes, perceived risks, and public health measures. This improvement facilitates more reliable characterization of public discourse, which is critical for supporting social media–based public health surveillance and informing communication strategies.
**Conclusions:** TM-Rephrase enhances the reliability and interpretability of topic modeling for short, noisy social media data in public health contexts, enabling more actionable insights for public health infoveillance. By improving semantic clarity without modifying underlying models, this framework provides a scalable and practical technique for monitoring public discourse and supporting evidence-based public health communication and decision-making.

# Introduction

### Background

The wide adoption of social media has fundamentally transformed public discourse, particularly during public health emergencies (e.g., the COVID-19 pandemic) [1]. Twitter (now X), characterized by rapid information dissemination and concise conversational format, has emerged as a critical yet methodologically challenging data source for understanding and analyzing public discussion topics and the dynamics of information exchange during the COVID-19 pandemic [2]. In addition to enabling real-time communication, these platforms support large-scale public engagement, where individuals actively express opinions, share experiences, and respond to official health communications. For public health agencies such as the U.S. Centers for Disease Control and Prevention (CDC), analyzing short-text messages in the format of public replies, opinions, and concerns provides valuable insights for informing more effective communication strategies to the public [3]. Such innovative surveillance and analyses can help capture evolving public concerns, assess responses to current health policies, and improve the alignment between institutional messaging and public needs. However, from a public health perspective, effectively leveraging these data remains challenging due to the unstructured nature of user-generated social media short texts, which complicate systematic monitoring and timely interpretation of population-level responses. In particular, the inherent brevity, informality, and ambiguity in individual posts, which is limited to 280 characters [4], pose significant challenges for traditional social media data mining and analytics, including topic modeling that discovers latent thematic structures of textual corpora, as the limited textual context constrains reliable semantic interpretation.

The contextual sparsity of social media short texts often leads to topics of keywords that are less coherent and less diverse (i.e., more overlapping topics with redundant or unrelated keywords), thereby reducing interpretability and limiting their utility for understanding various public opinions during the pandemic. This limitation is further compounded by the heterogeneous nature of online public discourse, where variations in writing styles and expression introduce additional complexity into textual data. This problem becomes particularly pronounced when attempting to capture emerging attitudes and thematic structures in communications of public health emergencies, where semantic ambiguity, brevity and informal linguistic styles (e.g., abbreviations, slang, layman expressions and misspellings) are prevalent [5]. Moreover, even when topic models identify clusters of topic keywords, they often lack meaningful interpretability without additional contextualization or external knowledge from experts, making it difficult to translate model outputs into coherent narratives for public health decision support. Consequently, these challenges limit the effectiveness of topic modeling in supporting systematic analysis of public health discourse. These issues suggest a need to standardize and formalize short social media texts into more context-formal content prior to application of topic modeling in health communications, thereby improving semantic clarity and enhancing the interpretability of downstream analytical results.

Topic modeling, as a widely used method for discovering latent thematic structures in textual corpora, is exemplified by the classic probabilistic model, Latent Dirichlet Allocation (LDA) [6], proposed by Blei et al. in 2003. However, despite its broad adoption, applying LDA or its variants to short texts from social media platforms poses substantial challenges. The brevity, informality, and noise of posts weaken statistical signals and exacerbate the context sparsity, leading to incoherent, redundant, or less interpretable topics generated from these probabilistic models [7].

To address these challenges, some enrichment strategies have been proposed. One line of research augments short texts with external knowledge or auxiliary contextual signals to enhance statistical robustness [8]. Another direction leverages distributed representations of words and documents, incorporating embeddings or neural architectures to map texts into richer semantic spaces. Prominent examples include BERTopic [9] and Top2Vec [10], which utilize more recent transformer-based embeddings and clustering methods to generate more semantically meaningful topics. Recent work has explored more advanced neural network architectures, such as FASTopic [11] and topic-semantic contrastive learning [12], which demonstrate improved handling on context sparsity and has the potential to be applied in health communications.

All these approaches focus on algorithmic model redesign with parameterized inference strategies for short texts, which face challenges on poor adaptation to real-life large-scale data (e.g., public health related discourse on social media). In addition, such topic modeling algorithms typically output sets of top-ranked keywords that can be difficult for humans without additional domain knowledge equipped to interpret.

Regarding the context of public discourse during the COVID-19, the pandemic fundamentally transformed social media into a primary platform for real-time public discourse, where X has been extensively used to study public attitudes and evolving concerns towards the pandemic and various health policies [13, 14]. Scholars have analyzed public reactions to official communications from health agencies such as the CDC, focusing on opinion and thematic feedback to health policies [3, 15]. Additional research has explored geospatial and demographic variations in public concerns, demonstrating the heterogeneous impact of the pandemic on different communities [16]. Despite the active research and the value of social media for effective public health infoveillance during emergencies , they often overlook the interpretability issues arising from short and noisy texts in online discourse especially on social media platforms.

Motivated by Maini et al. [17], we leverage large language models (LLMs) for rephrasing short texts for topic modeling, aiming to standardize and formalize public replies to CDC X posts by rephrasing these replies. This framework, termed *TM-Rephrase*, builds on recent advances in LLMs to achieve strong performance across diverse text processing capability, including text summarization [18], sentiment analysis [19], and information retrieval [20].

In this study, we examine the impact of TM-Rephrase on the quality of topics derived from several representative topic models, using a dataset of public replies to CDC X posts during COVID-19 pandemic as a case study. We hypothesize that rephrasing original short text public replies on social media through TM-Rephrase enhances topic quality in terms of both intra-topics semantic coherence and inter-topic diversity of topic model outputs, which are represented as sets of keywords. Specifically, we investigate two rephrasing schemes: (i) general rephrasing and (ii) colloquial-to-formal rephrasing that aim to align with official public health communications (e.g., from CDC), and evaluate their effectiveness across multiple topic modeling algorithms, including probabilistic algorithms, e.g. LDA [6] and more advanced neural network-based topic models (e.g., BERTopic [9] and FASTopic [11]).

**Research Questions and Objectives**

The research questions of this study are: (1) whether and how effectively TM-Rephrase improves the quality and interpretability of topic model outputs for short texts in health communications, (2) how different rephrasing schemes influence topic model performance, and (3) based on the topics, how to interpret public discussions as replies to CDC during the COVID-19 pandemic .

This study aims to improve the analysis of public responses to official public health communications on social media to support more effective public health surveillance. To achieve this, we develop and evaluate TM-Rephrase, an LLM–based model-agnostic framework that enhances the quality and interpretability of topic modeling for short-text public replies to CDC communications during the COVID-19 pandemic.

# Methods

TM-Rephrase is designed as a multi-stage pipeline to improve topic modeling performance in public health communications, as is shown in Figure 1. The pipeline is composed of four major stages: (1) data collection, (2) LLM-based rephrasing, (3) data preprocessing, and (4) topic modeling with a selected algorithm and evaluation of the resulting topics, as detailed in the following subsections. We will use this pipeline to study how rephrased text of public responses to CDC posts on X impacts topic modeling outcomes compared to non-rephrased benchmarks.

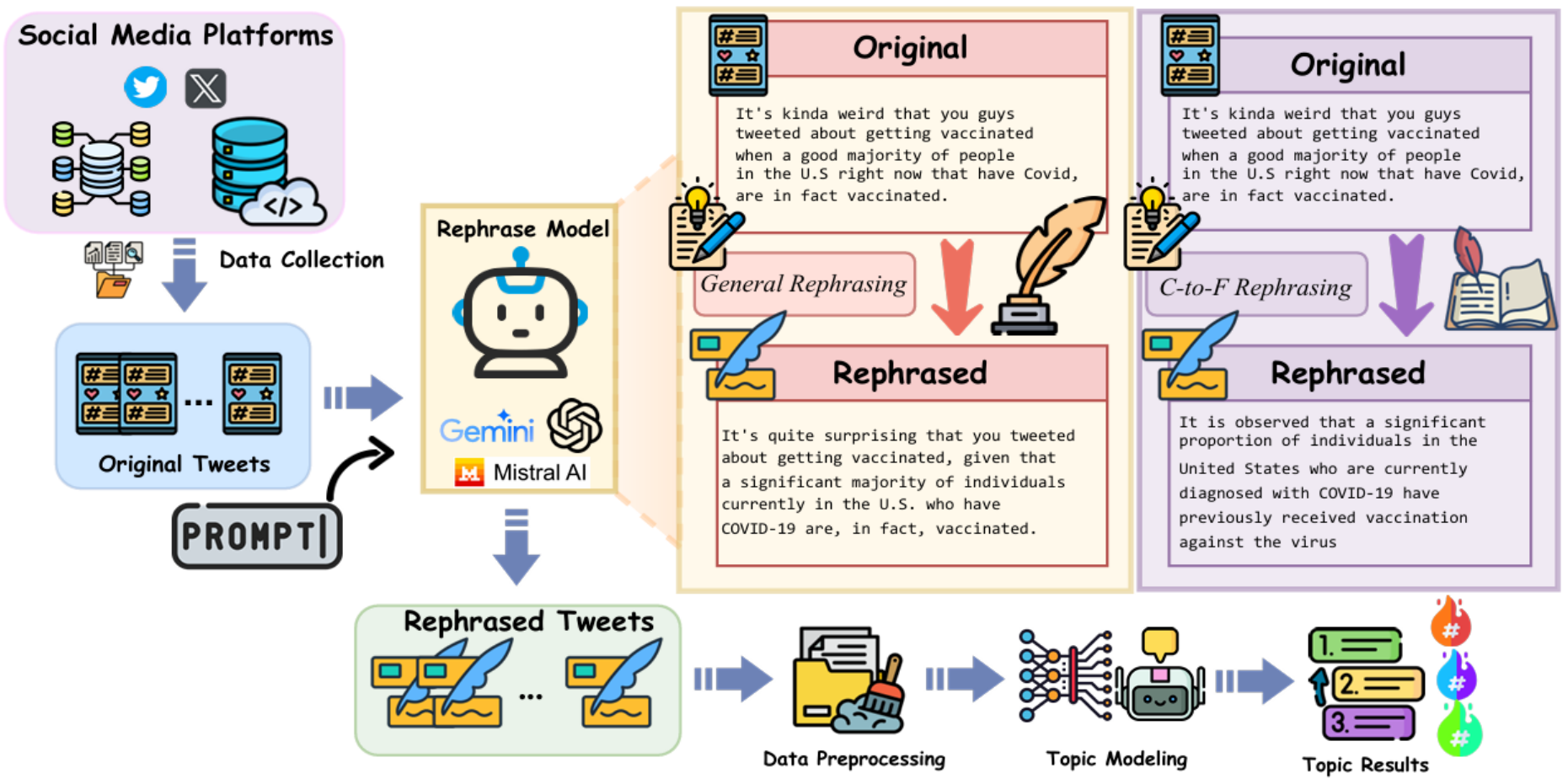

**Figure 1**. Overview of the TM-Rephrase framework. The pipeline integrates data collection, LLM-based rephrasing, data preprocessing, and topic modeling algorithm application with evaluation.

Before introducing the model details, we begin by formulating the problem setting with preliminary notations that are used throughout the paper. Let $D = \{d_1, \ d_2, \dots, d_N\}$ denote a corpus of *N* short-text tweets (i.e., replies to CDC posts). Each document $d_i$ is typically brief and context-limited, as the inherent sparsity characteristic of social media discourse.

The goal of topic modeling in this setting is to extract and model a set of *K* latent topics, denoted by $T = \{T_1, T_2, \dots, T_K\}$, where each topic $T_k$ is represented by a set of its top-*T* representative keywords, extracted or generated by the corresponding topic modeling algorithm, i.e., $T_k = \{w_1, w_2, \dots, w_T\}, k = 1, 2, \dots, K$. These representative keywords serve as the semantic expression of the topic and form the basis for downstream topic interpretation, evaluation, and further analysis.

**Data Collection**

Our study constructed and utilized a dataset of X posts that are public responses to CDC official communications during the COVID-19 pandemic (May 2020– November 2022). Data collection tracked seven official CDC accounts: @CDCgov, @CDCDirector, @CDCGlobal, @CDCMMWR, @CDCtravel, @DrReddCDC, and @CDCEmergency. Initial acquisition yielded 180,090 public interactions with official communications, including replies and mentions. A rigorous mapping process then linked these interactions to their original CDC official tweets, establishing conversational context. Filtering excluded approximately 49,000 direct mentions from private accounts (e.g., user @XXXX mentions @CDCEmergency from his account, which should not count as a reply) and 104,000 replies unlinked due to deleted or inaccessible original tweets. This resulted in a final dataset of 25,027 replies linked to 1,512 unique original CDC tweets. These reply posts capture a wide range of public opinions, concerns, and discussions surrounding official health communications during a major public health emergency. Statistics of these replies are presented in Table 1.

**Table 1.** Statistics of the Original and Rephrased Datasets of Public Replies to CDC Communications during the COVID-19 Pandemic.

| Quantity of Original and Rephrased Datasets | Original | General Rephrased | C-to-F Rephrased |
|---|---|---|---|
| Mean # of Words | 27.33 | 27.76 | 28.14 |
| Standard Deviation of Word Counts | 14.23 | 15.12 | 14.11 |
| Minimum # of Words | 1 | 1 | 1 |
| 25th Percentile | 15 | 16 | 16 |
| Median # of Words | 27 | 27 | 29 |
| 75th Percentile | 40 | 41 | 42 |
| Maximum # of Words | 64 | 66 | 68 |

Formally, the dataset $D = \{d_1,\ d_2, \dots, d_N\}$ denotes the original corpus, where each short-text reply, is denoted as:

$$d_i = \left(w_1^i, w_2^i, \dots, w_{L_i}^i\right), i = 1,2, \dots, N \quad (1)$$

which contains a limited number of tokens $L_i$ of post $d_i$ since each tweet reply is limited to 280 characters [4]. Such short and informal nature of these short texts results in sparse lexical co-occurrence signals, motivating the need for semantic enrichment prior to topic modeling algorithms applied.

## LLM-based Rephrasing

The second stage introduces an LLM-based rephrasing operator that transforms each short-text document (i.e., a piece of public reply to a CDC post) into a semantically and contextually refined version that is similar to CDC's official communication. Let $R_\theta$ denote a rephrase model (i.e., an LLM) parameterized by $\theta$. For each short-text reply $d_i$, the rephrasing process is defined as:

$$\widehat{d_\iota} = R_\theta\ (\pi, d_i), i = 1,2, \dots, N \quad (2)$$

where $\widehat{d_\iota}$ refers to the rephrased short-text reply and π denotes a carefully designed prompt that constrains the generation process to preserve semantic fidelity and ensure rephrasing executed under a certain scheme that will be introduced later. This rephrasing stage is crucial for the downstream topic modeling of the public replies and for better understanding of public responses to official health communications during emergencies.

For the rephrase stage, we experimented with three different LLMs: Google Gemini (gemini-2.5-flash [21]), OpenAI GPT (GPT-4o-mini [22]), and an open-sourced LLM (Mistral-7B-Instruct [23]), given their strong natural language generation capabilities and complementary trade-offs between performance and cost for large-scale applications. We used a low sampling temperature (0.2) to ensure semantic fidelity and minimize stochastic variation during automated rephrasing, consistent with findings from prior work [24].

To systematically investigate how different styles of linguistic refinement influence topic modeling performance, TM-Rephrase incorporates two distinct prompt-guided rephrasing schemes that we developed: *General Rephrasing* and *Colloquial-to-Formal (C-to-F) Rephrasing*. Although both schemes are designed to keep semantic fidelity, they differ in the degree and style of linguistic transformation applied to the original short texts of public replies. Prompts for these two schemes are shown in Supplemental Information A. Table 5.

*General Rephrasing:* The general rephrasing scheme performs light-weight linguistic refinement while strictly maintaining the meaning and informational content of the original text. Its primary objective is to improve grammatical correctness, syntactic clarity, and overall readability without altering domain-specific terminology, named entities, hashtags, usernames, or technical details. Concretely, the general phrasing scheme fixes typographical errors, incomplete sentence fragments, inconsistent verb tenses, and ambiguous phrasing, all of which are common in social media discourses. It may restructure fragmented clauses into complete sentences and resolve minor ambiguities that arise from informal writing conventions. However, it avoids introducing new concepts, removing key terms, or substituting specialized vocabulary. As such, general rephrasing can be viewed as a minimal-intervention normalization process that enhances structural regularity while preserving the lexical distribution of domain-relevant tokens.

*C-to-F Rephrasing:* The C-to-F rephrasing scheme transforms informal, conversational, or slang-based expressions into formal, professional English suitable for public health communication contexts. The rephrased outputs are structured to resemble the tone and style of public health reports or professional summaries, using complete sentences and standardized grammar while avoiding slang, contractions, and casual phrasing. During this process, implicit references—such as omitted subjects, abbreviated expressions, or context-dependent meanings (eg, “this shot,” “they said,” or fragmentary statements lacking clear referents)—are made explicit by clarifying the intended subject, action, or relationship within the sentence. Fragmented or elliptical constructions are rewritten into fully articulated and logically coherent statements to improve readability and interpretability. Strict constraints are applied to preserve the original meaning, including retaining all named entities, hashtags, usernames, and domain-specific terminology. No additional claims, interpretations, or external information are introduced.

These two rephrasing schemes enable a controlled comparison between moderate linguistic normalization and more substantial stylistic formalization for health communications. This design allows us to evaluate whether incremental grammatical refinement alone suffices to improve topic modeling performance, or deeper formalization is required for optimal topic quality. In order to ensure semantic alignment and maintain between the original and rephrased texts, the prompt explicitly instructs the LLMs to retain meaning and avoid modifying domain-specific content.

The resulting rephrased corpus is denoted as (3) where $\widehat{d_i}$ represent each individual rephrased short-text reply:

$$\widehat{D} = \{\widehat{d_1}, \widehat{d_2}, \ldots, \widehat{d_N}\} \quad (3)$$

**Data Preprocessing**

Before applying topic modeling, both the original and rephrased corpora undergo a consistent and standardized preprocessing procedure to ensure comparability and reduce extraneous noise. The preprocessing step includes removal of URLs, punctuation marks, adding stop words, which do not contribute semantic information to topic inference during topic modeling.

At the same time, semantically meaningful elements are carefully preserved. Emojis are retained through their textual descriptions to maintain their affective or contextual signals. For example, the emoji 😊 is converted to a textual representation ‘smiling face’, and 😷 is mapped to ‘face with medical mask’. This conversion ensures that non-textual, yet informative elements remain encoded within the lexical space rather than discarded as noise.

All texts are further normalized by converting characters to lowercase, followed by tokenization and lemmatization. Lowercasing eliminates case-based lexical duplication, tokenization segments text into analyzable units, and lemmatization reduces inflection forms of tokens to their base form, thereby consolidating semantically equivalent variants into unified lexical representations. All these operations promote lexical consistency and stabilize downstream word co-occurrence statistics in topic modeling.

Formally, let $P_\gamma(\cdot)$ denote the preprocessing operator parameterized by γ, which encapsulates the sequence of normalization, tokenization, and lemmatization in data preprocessing step. After preprocessing, each original document $d_i$ and rephrased document $\widehat{d_i}$ are respectively converted into the preprocessed format:

$$p_i = P_\gamma(d_i), \qquad \widehat{p_i} = P_\gamma(\widehat{d_i}), \qquad i = 1,2, \ldots, N \quad (4)$$

Here $p_i \in R^{|V|}$ and $\widehat{p_i} \in R^{|V|}$ denote the tokenized representations of the rephrased and original documents (i.e., short-text replies), respectively, under a shared vocabulary space $V$. Maintaining identical preprocessing across both corpora ensures that any observed differences in topic modeling performance can be attributed to the rephrasing rather than discrepancies in preprocessing or representation learning.

**Topic Modeling**

In the final stage of the pipeline, topic modeling algorithms are applied, and their performance is systematically evaluated. As illustrated in Figure 2, both the original and rephrased replies are modeled by the same topic models, and the resulting topics are compared and evaluated through quantitative metrics and qualitative evaluation. The original corpus $D$ and rephrased corpus $\widehat{D}$ are independently input into the same specific topic modeling algorithm $M_{TM}$ to ensure methodological consistency:

$$T = M_{TM}(D), \qquad \hat{T} = M_{TM}(\hat{D}) \quad (5)$$

This parallel modeling design yields two corresponding sets of topic outputs, i.e., one derived from the original texts and the other from the rephrased texts. This enables a direct and controlled comparison. Because of the same topic modeling procedure, hyperparameter settings for topic models and preprocessing configurations are maintained across both corpora. Therefore, any observed differences in topic quality can be credited to the rephrasing scheme rather than the specific topic modeling algorithm.

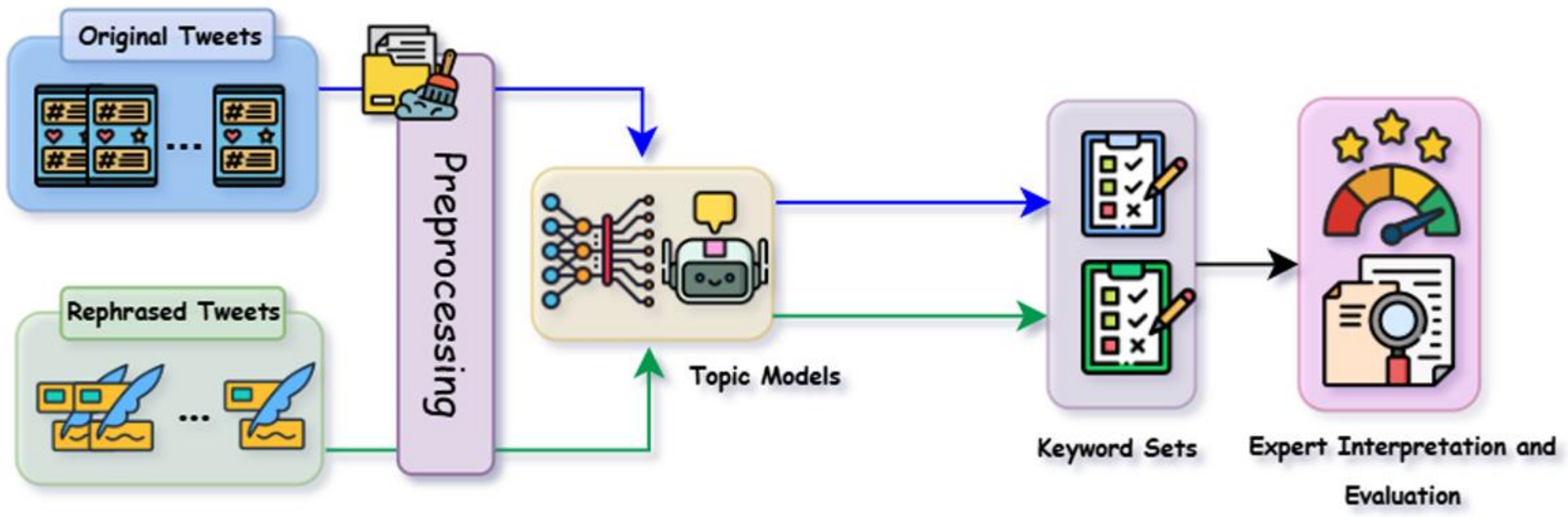

**Figure 2**. Outline of the topic evaluation process. Both the original and rephrased tweets were modeled using the same topic models, and the resulting topics were compared and evaluated through quantitative metrics and qualitative evaluation.

To comprehensively evaluate the effects of TM-Rephrase, three representative topic models were applied, including conventional statistical and neural network-based topic models. This suite of different topic models enables a systematic analysis of how different topic modeling algorithms respond to TM-Rephrase

**LDA** [6]: A foundational probabilistic topic model serving as the benchmark in topic modeling studies. It is effective for classic topic discovery but usually challenged by data sparsity in short social media texts.
**BERTopic** [9]: A neural topic model using contextual embeddings (e.g., Sentence-BERT [25]) and density-based clustering. It is considered well-suited for noisy, informal social media data.
**FASTopic** [11]: A ***F***ast, ***A***daptive, ***S***table, and ***T***ransferable topic model based on a transformer backbone that directly captures semantic relationships between document embeddings and learnable topic–word embeddings. It uses an embedding transport plan as an optimization objective to enhance topic–word and document–topic associations.
**TSCTM** [12]: As a ***T***opic-***S***emantic ***C***ontrastive ***T***opic ***M***odel, it addresses short-text sparsity via contrastive learning, generating dense semantic vectorized representations and topic distributions.

Together, these different topic models ensure that the evaluation of TM-Rephrase is not limited to a single topic modeling paradigm, but captures methodological robustness across multiple topic models. For all these four topic models, we fixed the number of topics ($K$) at 8, based on an exploratory analysis to balance between thematic detail and interpretability for the study dataset. For each topic, which is represented as a set of keywords, the top 15 keywords ($T = 15$) were set and ranked by the probability of model output.

The evaluation framework incorporates both quantitative and qualitative components, with primary emphasis on quantitative assessment. Specifically, topic coherence, topic uniqueness, topic redundancy, and topic diversity metrics are computed to measure both intra-topic internal semantic consistency, and inter-topic distinctiveness. These metrics provide objective indicators on whether and how semantic enhancement through rephrasing strengthens topic interpretability. Details on these quantitative metrics are described in the following section.

**Topic Modeling Performance Evaluation**

Four quantitative evaluation metrics are computed to quantify topic coherence, uniqueness, redundancy, and diversity, with details shown below.

*Topic Coherence ($C_v$ )*

$C_v$ [26] measures the semantic coherence among the topic keywords which are determined by topic-word probabilities. This metric was chosen based on the mathematical analysis and justifications in [27]. This metric fundamentally relies on Normalized Point wise Mutual Information (NPMI) to quantify pair-wise word semantic associations. For any two words $w_i$ and $w_j$, their NPMI score is defined as:

$$NPMI(w_i, w_j) = \frac{log\left(\frac{P(w_i, w_j) + \epsilon}{P(w_i)P(w_j) + \epsilon}\right)}{-log(P(w_i, w_j) + \epsilon)} \quad (6)$$

Here, $P(w_i)$ and $P(w_j)$ are the probabilities of observing words $w_i$ and $w_j$ respectively, and $P(w_i, w_j)$ is the co-occurrence probability of the word pair $(w_i, w_j)$ within a defined external corpus (Wikipedia corpus was used in this study [28]). A small constant $\epsilon$ (e.g., $10^{-12}$) is added to avoid division by zero or taking the logarithm of zero. Building on the NPMI formulation, the $C_v$ coherence score can be formally expressed as follows:

$$C_v = \frac{1}{T}\sum_{i=1}^{T} \cos\left(\mathbf{v}_{NPMI}(x_i), \mathbf{v}_{NPMI}\left(\{x_j\}_{j=1}^{T}\right)\right) \quad (7)$$

Here, $v_{NPMI}(x_i)$ represents the vector of NPMI scores between a word $x_i$ and all other words in the topic keyword list, defined as:

$$v_{NPMI}(x_i) = \{NPMI(x_i, x_j)\}_{j=1,\dots,T} \quad (8)$$

Similarly, the aggregated NPMI vector across all words in the topic keyword list is defined as:

$$\boldsymbol{v}_{NPMI}\left(\{x_j\}_{j=1}^{T}\right) = \left\{\sum_{i=1}^{T} NPMI(x_i, x_j)\right\}_{j=1,\dots,T} \quad (9)$$

where *T* is the number of top keywords in the topic (*T* = 15 in this study). The cosine similarity is then computed between the word-specific and aggregated vectors, and the final $C_v$ score is obtained by averaging across all words. A higher $C_v$ score indicates stronger semantic coherence and better interpretability of the topics.

*Topic Uniqueness (TU)*

Nan et al. [29] proposed *TU*, a metric that quantifies how distinct the top keywords within each topic are across the entire topic set $T$. This helps identify if individual topics leverage unique vocabulary effectively. Given *K* topics and the top *T* keywords of each topic, *TU* is computed as:

$$TU = \frac{1}{K}\sum_{k=1}^{K}\left(\frac{1}{T}\sum_{x_i \in T_k}\frac{1}{\#(x_i)}\right) \quad (10)$$

where $T_k$ denotes the top keyword set of the *k*-th topic, and $\#(x_i)$ denotes the occurrence count of word $x_i$ in the top *T* words of all topics. TU ranges from *1/K* to 1, and a higher *TU* score indicates more unique topics.

*Topic Redundancy (TR)*

*TR* [30] was developed to measure the overlap of top keywords across different topics. The formula for *TR* is given as:

$$TR = \frac{1}{K}\sum_{k=1}^{K}\left(\frac{1}{T}\sum_{x_i \in T_k}\frac{\#(x_i) - 1}{K - 1}\right) \quad (11)$$

A lower *TR* score suggests that the topics are more distinct.

*Topic Diversity (TD)*

The formula for *TD* [31], which measures the proportion of unique top keywords across all topics, is defined as:

$$TD = \frac{1}{K}\sum_{k=1}^{K}\frac{1}{T}\sum_{x_i \in T_k}\mathbb{I}\big(\#(x_i)\big) \quad (12)$$

A higher *TD* score indicates more diverse topics with less word overlap. This metric works by identifying words that are unique to a single topic's top keyword list. The indicator function $I(\cdot)$ is defined by a stepwise function. This function returns 1 if and only if a word $x_i$ appears in the top keyword list of exactly one topic, and 0 otherwise.

Taken together, these four metrics, $C_v$, *TU*, *TR* and *TD*, evaluate topic quality across complementary dimensions. In brief, $C_v$ measures semantic coherence among a topic top keywords, *TU* assesses topic uniqueness across topic, *TR* quantifies topic redundancy due to topical overlap, and *TD* evaluates whether the model produces a diverse set of topics. These metrics provide a comprehensive and systematic measure to evaluate and compare topics generated from original non-rephrased posts and those rephrased through two different TM-Rephrase schemes (general and C-to-F rephrasing).

In addition to quantitative assessment, qualitative evaluation is conducted through human expert assessment of interpretability and thematic alignment. This external evaluation examines whether the topics generated from rephrased texts exhibit clearer conceptual boundaries, improved readability, and stronger alignment with domain-relevant posts, while ensuring that rephrasing does not distort the original semantic meaning.

Through systematic comparisons ($T$ and $\hat{T}$), we evaluate the extent to which the model-agnostic TM-Rephrase framework improves the intra-topic coherence, inter-topic uniqueness, semantic robustness, and interpretability of topic assignments.

# Results

This section presents the results of quantitative evaluation of TM-Rephrase followed by qualitative assessments. We report results across four topic models, FASTopic, BERTopic, TSCTM, and LDA, under three conditions: (1) original tweets without any rephrasing, (2) general rephrasing, and (3) C-to F rephrasing, and using three different LLMs for rephrasing (i.e., Gemini-2.5-flash, GPT-4o-mini, and Mistral-7B Instruct). Topic quality was quantitatively assessed using four metrics: $C_v$, *TU*, *TR*, and *TD* as described the Method: Topic Modeling Performance Evaluation section.

Before introducing the quantitative results, we refer to Table 1 to assess potential text expansion or compression caused by TM-Rephrase. Both general and C-to-F rephrasing schemes hardly change token lengths: the mean increases marginally from 27.33 (original) to 27.76 (general) and 28.14 (C-to-F), and the median changes slightly from 27 to 29 (C-to-F). Quartiles and maxima show similar minor differences, indicating that rephrasing preserves the overall text length distribution, and any effects on topic modeling are unlikely to stem from token counts.

### Quantitative Results

Table 2 reports the quantitative evaluation results under a three-factor design, including four topic modeling algorithms, three LLM backends, and three rephrasing schemes, resulting in a total of 36 configurations. This design enables a systematic assessment of the main effects of rephrasing, as well as its interaction with model architecture and LLM choice.

*Effect of Rephrasing Schemes*

Across all configurations, TM-Rephrase consistently improves topic quality, with the most pronounced and stable gains observed in topic coherence ($C_v$). In particular, the colloquial-to-formal (C-to-F) rephrasing scheme generally yields the highest coherence scores. For example, with LDA topic model, coherence increases substantially from 0.3094 (without rephrasing) to 0.5004 under Gemini-2.5-flash, with similarly large improvements observed for GPT-4o-mini (0.4917) and Mistral-7B-Instruct (0.4762). These results indicate that rephrasing effectively mitigates contextual sparsity in short texts and enhances semantic consistency across topics.

**Table 2**. Quantitative metric results for various topic models, both with and without TM-Rephrase by three large language models (LLMs). [a-d]

| | | $C_v$(0–1) ↑ | *TU* (0.125–1) ↑ | *TR* (0–1) ↓ | *TD* (0–1) ↑ |
|---|---|---|---|---|---|
| | FASTopic w/o *rephr.* | .3388 | .9917 | .0024 | .975 |
| **Gemini-2.5-flash** | FASTopic w/ *general rephr.* | .3723 | .9917 | .0024 | .9917 |
| | FASTopic w/ *c-to-f rephr.* | .3301 | **1** | **0** | **1** |
| **GPT-4o-mini** | FASTopic w/ *general rephr.* | **.3746** | .9917 | .0024 | .9917 |
| | FASTopic w/ *c-to-f rephr.* | .3422 | **1** | **0** | **1** |
| **Mistral-7B-instruct** | FASTopic w/ *general rephr.* | .3728 | .9917 | .0024 | .9917 |
| | FASTopic w/ *c-to-f rephr.* | .3321 | .9917 | .0024 | .9917 |
| | BERTopic w/o *rephr.* | .4078 | .4667 | .3976 | .4667 |
| **Gemini-2.5-flash** | BERTopic w/ *general rephr.* | .4564 | **.525** | **.3357** | **.525** |
| | BERTopic w/ *c-to-f rephr.* | **.4734** | **.525** | .3452 | **.525** |
| **GPT-4o-mini** | BERTopic w/ *general rephr.* | .4612 | .4667 | **.3357** | .4667 |
| | BERTopic w/ *c-to-f rephr.* | .4687 | **.525** | .3452 | **.525** |
| **Mistral-7B-instruct** | BERTopic w/ *general rephr.* | .4574 | .4667 | **.3357** | .4667 |
| | BERTopic w/ *c-to-f rephr.* | .4713 | .4667 | .3976 | .4667 |
| | TSCTM w/o *rephr.* | .3094 | .9917 | .0024 | .975 |
| **Gemini-2.5-flash** | TSCTM w/ *general rephr.* | .3145 | .9833 | .0048 | .9833 |
| | TSCTM w/ *c-to-f rephr.* | **.3394** | **1** | **0** | **1** |
| **GPT-4o-mini** | TSCTM w/ *general rephr.* | .3267 | .9917 | .0024 | .9917 |
| | TSCTM w/ *c-to-f rephr.* | .3331 | **1** | **0** | **1** |
| **Mistral-7B-instruct** | TSCTM w/ *general rephr.* | .3104 | .9833 | .0024 | .9833 |
| | TSCTM w/ *c-to-f rephr.* | .3213 | .9917 | .0024 | .975 |
| | LDA w/o *rephr.* | .3094 | **.575** | .3095 | **.575** |
| **Gemini-2.5-flash** | LDA w/ *general rephr.* | .4206 | .5583 | .3214 | .5583 |
| | LDA w/ *c-to-f rephr.* | **.5004** | **.575** | **.3048** | **.575** |
| **GPT-4o-mini** | LDA w/ *general rephr.* | .4197 | .5583 | .3214 | .5583 |
| | LDA w/ *c-to-f rephr.* | .4917 | **.575** | .3095 | **.575** |
| **Mistral-7B-instruct** | LDA w/ *general rephr.* | .4008 | .5583 | .3095 | .5583 |
| | LDA w/ *c-to-f rephr.* | .4762 | .5583 | .3095 | .5583 |

[a]Table rows in light gray are results for TM-Rephrase based on Gemini-2.5-flash , light green for GPT-4o-mini, and light blue for Mistral-7B-Instruct.

[b]The best results within each model group are highlighted **in bold** and highlighted.

[c]Arrows in the header indicate the desired direction for each metric (higher ↑ or lower ↓ is better).

[d]*general rephr.* stands for the general rephrasing scheme while *c-to-f rephr.* stands for the colloquial-to-formal rephrasing scheme.

*Effect Across Topic Modeling Algorithms*

The magnitude of improvement varies by topic models. Probabilistic models such as LDA exhibit the largest gains, reflecting their strong dependence on lexical co-occurrence patterns.

BERTopic shows consistent but more moderate improvements, with coherence increasing across all LLMs and rephrasing schemes, and reaching its highest value under Gemini-2.5-flash ($C_v$ = 0.4734). In contrast, FASTopic demonstrates relatively smaller and less consistent gains, with slight decreases in coherence under the C-to-F scheme in some cases (e.g., $C_v$ = 0.3301 under Gemini-2.5-flash). This suggests that embedding-based models may already partially capture semantic relationships despite informal language, reducing the effectiveness of input standardization. TSCTM, meanwhile, benefits substantially from rephrasing in terms of diversity-related metrics, achieving perfect topic separation ($TU$ = 1, $TD$ = 1, $TR$ = 0) under C-to-F rephrasing with both Gemini-2.5-flash and GPT-4o-mini.

*Effect of LLM Backbones*

The improvements from TM-Rephrase are robust across all three LLM backbones. Although minor variations in absolute performance are observed, the patterns remain clear and consistent: rephrasing improves coherence and diversity metrics across topic models regardless of the underlying LLM. For example, LDA coherence gains and TSCTM diversity improvements are observed consistently across Gemini-2.5-flash, GPT-4o-mini, and Mistral-7B-Instruct. This stability suggests that the effectiveness of TM-Rephrase is not dependent on a specific LLM, but rather reflects a generalizable rephrasing strategy.

*Diversity and Redundancy Metrics*

In addition to more intra-topic coherence, rephrasing also improves inter-topic quality. Across FASTopic, BERTopic, and TSCTM topic models, topic uniqueness ($TU$) and diversity ($TD$) generally increase, while redundancy ($TR$) decreases or remains similar. The most notable improvement is observed in TSCTM under C-to-F rephrasing, where perfect diversity scores are achieved, indicating fully distinct topic clusters. These findings strongly suggest that rephrasing enhances not only intra-topic coherence but also inter-topic separation.

Figure 3 shows that C-to-F rephrasing consistently outperforms general rephrasing, while the overall improvement patterns remain highly consistent across Gemini-2.5-flash, GPT-4o-mini, and Mistral-7B-Instruct. These results demonstrate the robustness of the TM-Rephrase framework as well.

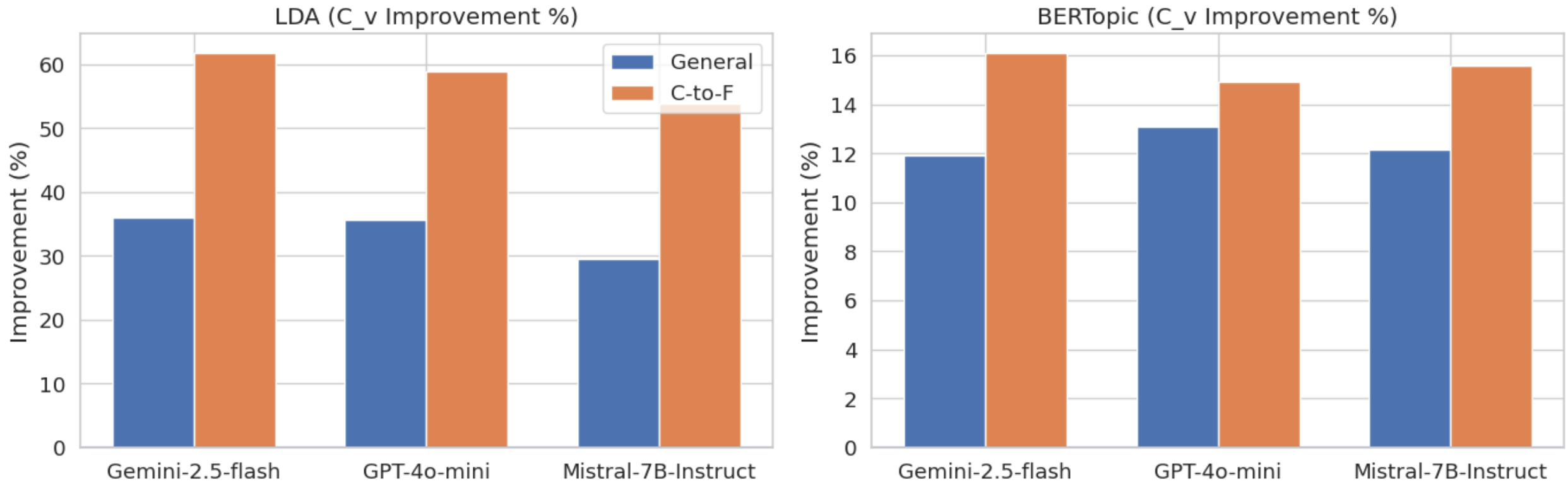

**Figure 3**. $C_v$ improvement (%) across LLMs and rephrasing schemes for LDA and BERTopic.

Besides, a nuanced trade-off between intra-topic coherence and inter-topic diversity emerges, especially for the C-to-F rephrasing scheme. While this scheme often delivers the highest coherence and perfect diversity scores, it can occasionally reduce topic uniqueness in certain topic models. For example, C-to-F rephrasing maximizes coherence across all LLMs but slightly lowers TU compared to the non-rephrased baseline in LDA. This suggests that aggressive formalization might homogenize lexical choices across topics, highlighting the importance of matching rephrasing strength with model characteristics. In addition, the comparison across LLM backbones reveals a high degree of robustness in TM-Rephrase. While Gemini 2.5-flash often achieves the strongest absolute results, GPT 4o-mini and Mistral-7B-Instruct closely track its performance across all metrics and models. The relative ranking of rephrasing schemes and topic models remains largely unchanged, indicating that TM-Rephrase is not tightly attributed to a specific LLM but instead constitutes a generalizable topic model-agnostic paradigm.

Together, these quantitative results provide strong evidence that TM-Rephrase consistently enhances intra-topic coherence and almost all inter-topic quality metrics for short, noisy texts in health communications across multiple topic models and LLM backbones. The two different rephrasing schemes that we developed have overall similar performance. The general rephrasing scheme offers balanced and reliable improvements, while the C-to-F scheme has a slight edge especially for lexically sensitive topic models such as LDA and BERTopic. These findings demonstrate effectiveness and robustness of our TM-Rephrase as an LLM-based model-agnostic framework for effective topic modeling in health communications.

**Qualitative Results**

Beyond quantitative improvements of rephrasing, qualitative evaluation also shows that rephrased posts yield more semantically coherent topics while reducing ambiguity in raw social media texts. Rephrasing is able to replace informal or loosely related terms with semantic, informative and more distinct keyword sets. As illustrated in Table 3, we shows all 8 topics by LDA without and with rephrasing (backboned on Gemini-2.5-flash).

In Table 3, LDA-derived topics without rephrasing exhibit limited semantic coherence, particularly for topic 2 (public health measures), which is dominated by generic and less informative terms (e.g., “mask,” “wear,” “stop”) and includes off-topic noise such as “dont.” General rephrasing improves topical specificity by introducing more contextually meaningful health-related terms (e.g., “personal_protective_equipment,” “social_distance”), although some colloquial artifacts persist (e.g., “please” and “could”). In contrast, C-to-F rephrasing produces more coherent and structured topics, characterized by standardized and policy-relevant terminology (e.g., “mandate,” “public_measure”). A similar pattern is observed in vaccine-related topics (e.g., topic 5), where baseline outputs contain vague terms (e.g., “like,” “get”), while rephrasing, particularly C-to-F, yields more precise and analytically meaningful keywords (e.g., “adverse,” “associated,” “efficiency”). Generally, increasing linguistic formality and contextual explicitness enhances topic coherence and reduces less informative terms, resulting in topics that express more specifically with professional public health discourse and support more reliable infoveillance.

We also assess the semantic alignment between replies’ content and the topics assigned to them by the topic models in Table 4.

As illustrated in Table 4, rephrasing improves the semantic alignment between tweet content and assigned topics across both LDA and BERTopic. In Example 1, the original tweet expresses concerns about vaccine side effects using informal and fragmented language, resulting in generic or weakly focused topic keywords (e.g., "vaccine," "child"). General rephrasing introduces explicit health-related terms such as "side effect," "risk," and "concern," improving thematic alignment. C-to-F rephrasing further strengthens this alignment by incorporating more formal and domain-specific terminology (e.g., "adverse"), enabling both models to produce more coherent and semantically grounded topics. A similar pattern is observed in Example 2, where the original tweet contains colloquial and emotionally charged expressions regarding Pfizer.

**Table 3.** Top 15 keywords for LDA topics based on Gemini-2.5-flash ($K$ = 8).[a, b]

| Topic ID | Keywords |
|---|---|
| | w/o *rephrasing* |
| 1 | vaccine, covid, get, people, dont, mask, kid, cdc, child, need, school, vaccinated, like, work, trump |
| 2 | mask, covid, wear, virus, vaccine, wearing, people, stop, dont, flu, need, face, work, social, spread |
| 3 | covid, death, vaccine, cdc, people, case, day, many, child, one, dont, symptom, week, died, year |
| 4 | vaccine, effect, dose, people, covid, know, covaxin, dont, pfizer, shot, like, mrna, child, long, one |
| 5 | covid, death, rate, case, cdc, data, infection, people, immunity, vaccination, study, show, variant, state, number |
| 6 | vaccine, cdc, covid, pfizer, stop, fda, people, transmission, please, health, year, child, approved, public, prevent |
| 7 | covid, ivermectin, dont, mask, cdc, people, work, vaccine, stop, positive, tested, like, getting, dose, doctor |
| 8 | vaccine, covid, vaccinated, get, flu, still, shot, people, child, year, got, even, booster, fully, getting |
| | w/ *general rephrasing scheme* |
| 1 | covid, cdc, vaccine, death, please, mask, pandemic, trump, virus, could, american, health, public, regarding, trust |
| 2 | mask, vaccinate, wear, covid, please, personal_protective_equipment, get, vaccine, school, individual, society, social_distance, virus, wearing, still |
| 3 | covid, vaccine, individual, flu, vaccinated, people, case, virus, immunity, child, stop, positive, tested, get, one |
| 4 | vaccine, covid, ivermectin, mrna, pfizer, people, lie, individual, death, effective, health, covaxin, received, moderna, treatment |
| 5 | vaccine, covid, child, shot, booster, effective, mrna, prevent, flu, please, receive, cdc, people, covaxin, variant |
| 6 | mask, covid, day, cases, test, people, virus, individual, cdc, one, work, vaccine, hospital, positive, wear |
| 7 | covid, child, vaccine, death, risk, vaccination, data, year, people, cdc, rate, individual, adverse, virus, cause |
| 8 | vaccine, risk, side_effect, testing, tweet, pfizer, would, covid, child, vaccination, kid, also, hospital, concern, original |
| | w/ *C-to-F rephrasing scheme* |
| 1 | mask, covid, individual, vaccine, public, operation, vaccinated, state, virus, health, wear, public_measure, may, work, mandate |
| 2 | covid, child, vaccine, individual, vaccination, regarding, death, concern, risk, year, virus, age, variant, may, significant |
| 3 | covid, individual, vaccine, positive, test, statement, author, mask, vaccination, testing, result, current, case, information, president |
| 4 | disease, cdc, prevention, control, center, covid, tweet, ivermectin, individual, health, professional, public, treatment, travel, coronavirus |
| 5 | vaccine, covid, effect, associated, data, symptom, adverse, cdc, child, efficiency, vaccination, side_effect, potential, disease, death |
| 6 | individual, vaccine, vaccination, covid, vaccinated, regarding, may, immunity, infection, health, efficacy, influenza, public, risk, concern |
| 7 | vaccine, vaccination, covid, individual, booster, pfizer, received, pharmaceutical, dose, trust, cdc, company, fda, financial, administration |
| 8 | public, health, regarding, covid, concern, pandemic, vaccine, social_distance, significant, information, current, mask, trump, cdc, individual |

[a]Irrelevant keywords (off-topic words that are decided by expert interpretation) are highlighted.
[b]Results with rephrasing are based on the input texts rephrased through Gemini-2.5-flash.

Topic assignments from the original text include noisy and less informative terms (e.g., "people," "dont," "get"). General rephrasing improves relevance by introducing clearer semantic cues (e.g., "concern"), while C-to-F rephrasing further refines the discourse with more structured and policy-relevant vocabulary (e.g., "trust," "responsibility," "profit"), allowing topic models, particularly BERTopic, to capture higher-level institutional themes.

**Table 4.** Examples of original, general, and C-to-F rephrased tweet replies with corresponding LDA and BERTopic topic assignment. [a, b]

| Text Type | Text Content | Topic Assigned by LDA | Topic Assigned by BERTopic |
|---|---|---|---|
| *Example 1.* | | | |
| *Original* | Are you kidding me with this? Don't be stupid enough to let your perfect baby be a guinea pig for these people. People are getting bells palsy after getting the vaccine. | **vaccine**, covid, get, people, dont, mask, **kid**, cdc, **child**, need, school, **vaccinated**, like, work, trump | covaxin, novavax, **vaccine**, approve, mrna, booster, ichoosecovaxin, need, want, fda, **safe**, people, **child**, wait, ocugen |
| *General Rephrased* | Are you serious? Don't be foolish and allow your healthy child to be a test subject for these individuals. Reports indicate people are developing Bell's palsy after receiving the vaccine. | **vaccine**, **risk**, **side_effect**, **testing**, tweet, pfizer, would, covid, **child**, **vaccination**, **kid**, also, hospital, **concern**, original | **vaccine**, covaxin, omicron, novavax, variant, mrna, delta, booster, approve, **child**, vaccinate, **side_effect**, people, **risk**, **concern** |
| *C-to-F Rephrased* | Concerns have been raised regarding potential adverse effects, specifically Bell's palsy, following vaccination. It is imperative to approach public health recommendations and medical interventions with informed decision-making. | **vaccine**, covid, effect, associated, data, symptom, **adverse**, cdc, **child**, efficiency, **vaccination**, **side_effect**, potential, disease, death | **vaccine**, covaxin, novavax, mrna, approval, **subject**, receive, **adverse**, booster, vaccination, **risk**, efficacy, **child**, **side_effect**, **concern** |
| *Example 2.* | | | |
| *Original* | Yes - let's trust Pfizer - same company that entered a $2.3 billion criminal plea deal for lying! Same company that is exempt from liability and CANT be sued for any injuries - yup sign me right up *eye roll* | **vaccine**, effect, dose, people, covid, know, covaxin, dont, **pfizer**, shot, like, mrna, child, long, one | **pfizer**, **vaccine**, effect, mrna, covid, cdc, moderna, people, shot, child, test, get, know, stop, receive |
| *General Rephrased* | Should we trust Pfizer? The same company that previously entered a $2.3 billion criminal plea deal for lying is now exempt from liability and cannot be sued for any injuries. So, yes, sign me right up. *eye roll* | **vaccine**, covid, ivermectin, mrna, **pfizer**, people, **lie**, individual, death, effective, health, covaxin, received, moderna, treatment | **vaccine**, **pfizer**, individual, vaccination, trial, mrna, **concern**, **lie**, report, utility, public, moderna, cdc, fda, information |
| *C-to-F Rephrased* | Concerns exist regarding public trust in Pfizer, given its history, including a $2.3 billion criminal plea deal for deceptive practices. Furthermore, the company's exemption from liability and inability to be sued for injuries raises significant questions regarding accountability. | **vaccine**, **vaccination**, covid, individual, booster, **pfizer**, received, pharmaceutical, dose, **trust**, cdc, **company**, fda, financial, administration | **vaccine**, **pfizer**, individual, **vaccination**, **responsibility**, mrna, **concern**, **profit**, business, **trust**, public, moderna, cdc, fda, information |

[a]Words that can better describe topics are **in bold** and highlighted.
[b]Results with rephrasing are based on the input texts rephrased through Gemini-2.5-flash.

# Discussion

## Principal Findings

In this study, we developed and evaluated TM-Rephrase, a model-agnostic framework that leverages advanced natural language generation capabilities of LLMs to rephrase original informal texts into standardized and grammatically sound expressions in public health contexts, while maintaining semantic fidelity. To ensure alignment with the original content, the rephrasing process is guided by carefully designed LLM prompts that emphasize both contextual and semantic fidelity. Using a large-scale dataset of 25,027 public replies to CDC posts on X during the COVID-19 pandemic, we demonstrated that rephrasing noticeably enhances both quantitative intra- and inter-topic quality and qualitative interpretability across multiple topic modeling algorithms.

**RQ1: Whether and how effectively TM-Rephrase improves topic quality and interpretability?**

Our findings demonstrate that TM-Rephrase substantially improves both quantitative topic quality and qualitative interpretability across all evaluated topic models. In particular, topic coherence consistently increases, with the largest gains observed in LDA (from 0.3094 to 0.5004, >60% relative improvement), indicating that rephrasing effectively mitigates contextual sparsity in short texts. Improvements are also observed in embedding-based models such as BERTopic, suggesting that enhanced linguistic clarity benefits a wide range of topic modeling approaches. In addition, inter-topic quality improves, as reflected by increased topic uniqueness and diversity and reduced redundancy. Qualitative evaluation further shows that rephrased texts yield more semantically aligned and interpretable topics, with clearer representation of public health themes. Together, these results indicate that TM-Rephrase improves both the internal coherence and practical interpretability of topic model outputs for short, noisy social media data.

**RQ2: How different rephrasing schemes influence topic model performance?**

The two rephrasing schemes exhibit distinct effects. General rephrasing provides consistent and balanced improvements by enhancing grammatical clarity while preserving original lexical structure. In contrast, the C-to-F scheme yields the strongest improvements in coherence and diversity metrics, particularly for lexically sensitive models such as LDA and TSCTM (e.g., achieving TU=1.0, TD=1.0, TR=0). However, more aggressive formalization may introduce minor trade-offs, such as slight reductions in topic uniqueness in some settings, suggesting that the degree of linguistic transformation should be aligned with model characteristics. Overall, these findings indicate that prompt design plays a critical role in shaping topic modeling performance, with C-to-F rephrasing offering the greatest benefits for enhancing semantic structure and interpretability.

**RQ3: How the resulting topics inform interpretation of public responses to CDC communications?**

The improved topic representations enable more meaningful interpretation of public discourse in response to CDC communications during the COVID-19 pandemic. The identified topics more clearly capture key public health concerns, including vaccination attitudes, perceived risks, and preventive measures. Compared to baseline results, rephrased topics reduce ambiguity and better align with the underlying semantics of user-generated content, allowing clearer identification of dominant themes and patterns in public responses. This improved interpretability supports more reliable characterization of population-level concerns and enhances the utility of social media data for public health infoveillance and communication analysis.

From a public health perspective, these findings highlight the importance of improving data quality as a foundational step in social media analytics. By transforming informal and noisy user-generated short texts into more structured and standardized content, TM-Rephrase enables more interpretable analysis of public discourse, thereby supporting more effective public health surveillance and communication strategies based on public discourse.

## Limitations and Future Work

This study still has several limitations that need to be addressed in future studies. First, the analysis focuses on public replies to official CDC communications on X during the COVID-19 pandemic. This specific setting provides a policy-relevant and well-defined context for infoveillance; however, patterns of engagement and discourse may differ across other platforms, health issues, and populations. Future research is needed to assess the generalizability of the proposed TM-Rephrase framework in broader contexts.

Second, LLM-based rephrasing process is designed to preserve semantic meaning while improving linguistic clarity. Although qualitative evaluation suggests that the core intent of original messages is largely maintained, minor variations in tone or emphasis may occur. In public health applications, where accurate interpretation of public concerns is critical, further validation using structured human evaluation or complementary quantitative measures should be deployed.

Last, the study design adopts a fixed number of topics and a selected set of representative topic modeling approaches to enable consistent and interpretable comparisons between rephrased and original topics. While this facilitates methodological evaluation, public health communication is inherently dynamic and context-dependent. Future work could adapt this framework to better reflect evolving public concerns during different stages of health emergencies.

## Conclusions

This study highlights the value of improving input text quality for online, especially social media–based public health infoveillance and analytics. By leveraging LLM-based rephrasing, TM-Rephrase enables more interpretable, consistent, and reliable detection of thematic patterns in short, noisy public responses to official health communications during health emergencies.

The data-centric and model-agnostic TM-Rephrase framework also offers a practical and scalable solution to enhance the usability of large-scale social media data for infoveillance. The findings suggest that linguistic standardization plays an important role in strengthening the interpretability of analytical outputs and supporting clearer understanding of public concerns from massive raw online data.

As social media continues to serve as a key channel for public engagement on various health topics, novel analytical frameworks, such as TM-Rephrase, enable more effective monitoring of public discourse and better-informed public health communication strategies, especially during health emergencies.

## Acknowledgements

This study is supported by NSF DMS-2436227.

## Conflicts of Interest

None declared.

## Abbreviations

BERT: Bidirectional Encoder Representations from Transformers
CDC: Centers for Disease Control and Prevention
COVID-19: Coronavirus Disease 2019
FASTopic: Fast, Adaptive, Stable, and Transferable topic model
GPT: Generative Pre-trained Transformer
LDA: latent Dirichlet allocation
LLMs: Large Language Models
TD: Topic Diversity
TR: Topic Redundancy
TSCTM: Topic-Semantic Contrastive Topic Model
TU: Topic Uniqueness